\documentclass[letterpaper, 10 pt, conference]{ieeeconf}  

\IEEEoverridecommandlockouts                              

\title{\LARGE \bf ADMM-Based Distributed MPC with Control Barrier Functions for Safe Multi-Robot Quadrupedal Locomotion}

\usepackage{amsmath,amssymb,amsfonts}
\usepackage{cite}
\usepackage{graphicx} 
\usepackage{epsfig} 
\usepackage{times} 
\usepackage{xcolor}
\usepackage{balance}
\usepackage{multicol}
\usepackage{array}
\usepackage{multirow}
\usepackage{booktabs}
\usepackage[colorlinks=true, allcolors=blue]{hyperref}

\newcommand{\Real}{\mathbb{R}}
\newcommand{\col}{\textrm{col}}
\newcommand{\Integer}{\mathbb{Z}_{\geq0}}
\newcommand{\diag}{\textrm{diag}}

\newcommand{\des}{\textrm{des}}

\newcommand{\terminal}{\textrm{terminal}}
\newcommand{\stage}{\textrm{stage}}

\newcommand{\thr}{\textrm{th}}
\newcommand{\Vset}{\mathcal{V}}
\newcommand{\Eset}{\mathcal{E}}
\newcommand{\Xset}{\mathcal{X}}
\newcommand{\Uset}{\mathcal{U}}
\newcommand{\Oset}{\mathcal{O}}
\newcommand{\Ggraph}{\mathcal{G}}
\newcommand{\Sset}{\mathcal{S}}
\newcommand{\Cset}{\mathcal{C}}
\newcommand{\Nset}{\mathcal{N}}

\DeclareMathOperator*{\argmin}{arg\,min}

\usepackage{etoolbox}

\makeatletter
\patchcmd{\@makecaption}{\scshape}{}{}{}
\makeatother

\AtBeginEnvironment{equation}{\vspace{-3pt}}
\AtEndEnvironment{equation}{\vspace{-3pt}}
\AtBeginEnvironment{align}{\vspace{-3pt}}
\AtEndEnvironment{align}{\vspace{-3pt}}
\AtBeginEnvironment{alignat}{\vspace{-3pt}}
\AtEndEnvironment{alignat}{\vspace{-3pt}}

\newtheorem{theorem}{\textbf{Theorem}}

\newtheorem{definition}{\textbf{Definition}}

\author{Yicheng Zeng$^{1}$, Ruturaj S. Sambhus$^{1}$, Basit Muhammad Imran$^{1}$, Jeeseop Kim$^{2}$, \\Vittorio Pastore$^{1}$, and Kaveh Akbari Hamed$^{1}$
\thanks{The work of R.S.~Sambhus and K.~Akbari Hamed is partially supported by the NSF under Grant 2423725.}
\thanks{$^{1}$Y.~Zeng, R.S.~Sambhus, B.~Imran, V.~Pastore, and K.~Akbari Hamed (\textit{Corresponding Author}) are with the Department of Mechanical Engineering, Virginia Tech, Blacksburg, VA 24061, USA, {\tt\small \{zyicheng, ruturajsambhus, basit, vittoriopastore, kavehakbarihamed\}@vt.edu}}
\thanks{$^{2}$J.~Kim is with The University of Texas at El Paso, El Paso, TX 79968, USA, {\tt\small jkim16@utep.edu}}
}

\begin{document}

\maketitle
\thispagestyle{empty}
\pagestyle{empty}


\begin{abstract}
This paper proposes a fully decentralized model predictive control (MPC) framework with control barrier function (CBF) constraints for safety-critical trajectory planning in multi-robot legged systems. The incorporation of CBF constraints introduces explicit inter-agent coupling, which prevents direct decomposition of the resulting optimal control problems. To address this challenge, we reformulate the centralized safety-critical MPC problem using a structured distributed optimization framework based on the alternating direction method of multipliers (ADMM). By introducing a novel node–edge splitting formulation with consensus constraints, the proposed approach decomposes the global problem into independent node-local and edge-local quadratic programs that can be solved in parallel using only neighbor-to-neighbor communication. This enables fully decentralized trajectory optimization with symmetric computational load across agents while preserving safety and dynamic feasibility. The proposed framework is integrated into a hierarchical locomotion control architecture for quadrupedal robots, combining high-level distributed trajectory planning, mid-level nonlinear MPC enforcing single rigid body dynamics, and low-level whole-body control enforcing full-order robot dynamics. The effectiveness of the proposed approach is demonstrated through hardware experiments on two Unitree Go2 quadrupedal robots and numerical simulations involving up to four robots navigating uncertain environments with rough terrain and external disturbances. The results show that the proposed distributed formulation achieves performance comparable to centralized MPC while reducing the average per-cycle planning time by up to 51\% in the four-agent case, enabling efficient real-time decentralized implementation.
\end{abstract}


\vspace{-0.3em}
\section{Introduction}
\label{sec:Introduction}

Quadrupedal robots have achieved significant progress in recent years, demonstrating robust locomotion over challenging terrain and enabling applications such as inspection, search and rescue, and collaborative transportation. Extending quadrupedal locomotion to multi-robot teams enables improved efficiency and robustness but introduces significant challenges due to high-dimensional dynamics, hybrid contact behaviors, and inter-agent safety constraints.

Model predictive control (MPC) has emerged as a powerful framework for trajectory generation and control of legged robots, as it enables systematic handling of system dynamics and constraints while optimizing performance objectives over a prediction horizon. MPC has been successfully applied using both reduced-order and full-order dynamic models of quadrupedal robots. Reduced-order template models capture the essential dynamics of locomotion while enabling computationally efficient trajectory optimization~\cite{Full_Koditschek_Template}. Common examples include the linear inverted pendulum (LIP) model~\cite{kajita19991LIP} and its extensions, such as the angular momentum LIP~\cite{ALIP}, spring-loaded inverted pendulum (SLIP)~\cite{SLIP}, vertical SLIP~\cite{vLIP_Sreenath}, and hybrid LIP~\cite{HLIP_Ames}, as well as centroidal dynamics models~\cite{orin2013centroidal} and the single rigid body (SRB) model~\cite{Kim_Wensing_Convex_MPC_01,Wensing_VBL_HJB,Abhishek_Hae-Won_TRO,Leila_Hamed_RAL,pandala2022robust,Ruturaj_Locomanipulation_RAL}. Alternatively, whole-body nonlinear MPC approaches leverage full-order robot dynamics to achieve higher accuracy and dynamic performance, albeit at increased computational cost~\cite{Patrick_TRO_Review,crocoddyl,ProxDDP_TRO}.


\begin{figure}
    \centering
    \includegraphics[width=\linewidth]{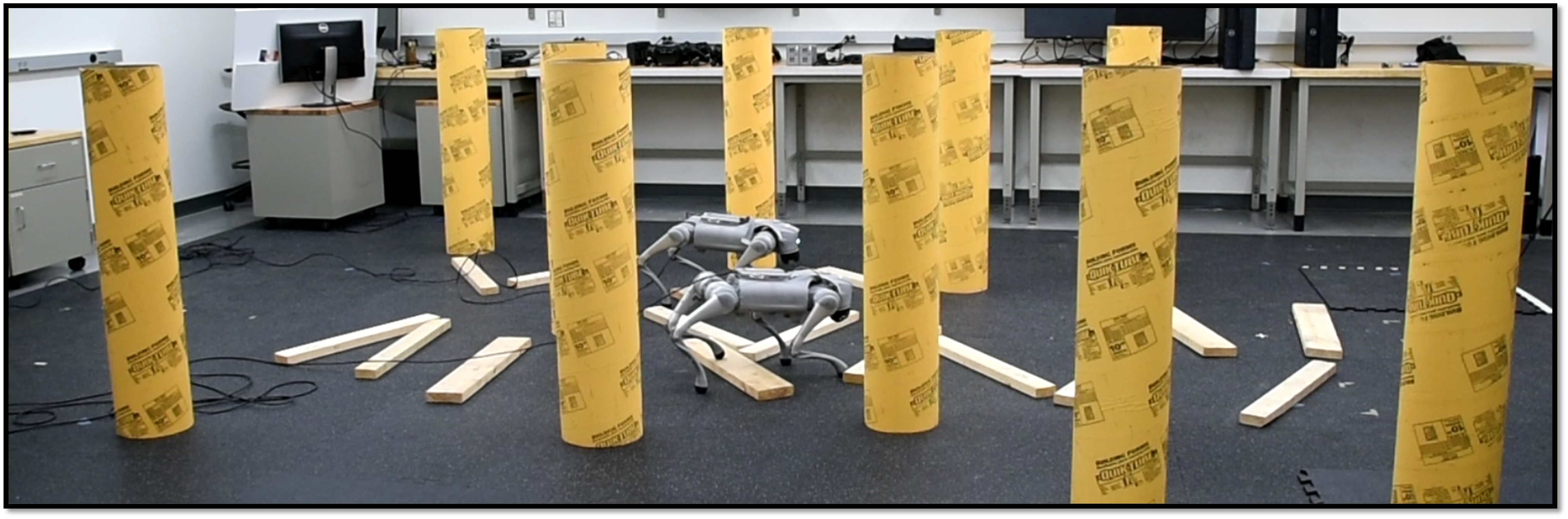}
    \vspace{-1.8em}
    \caption{Snapshot of the hardware experiment demonstrating the proposed ADMM-based CBF-DMPC framework, where two Unitree Go2 robots navigate a cluttered environment with rough terrain while maintaining safety-critical inter-agent and obstacle avoidance constraints.} 
    \vspace{-2.1em}
    \label{fig:OneSanpshot}
\end{figure}


\begin{figure*}[t]
    \centering
    \includegraphics[width=\linewidth]{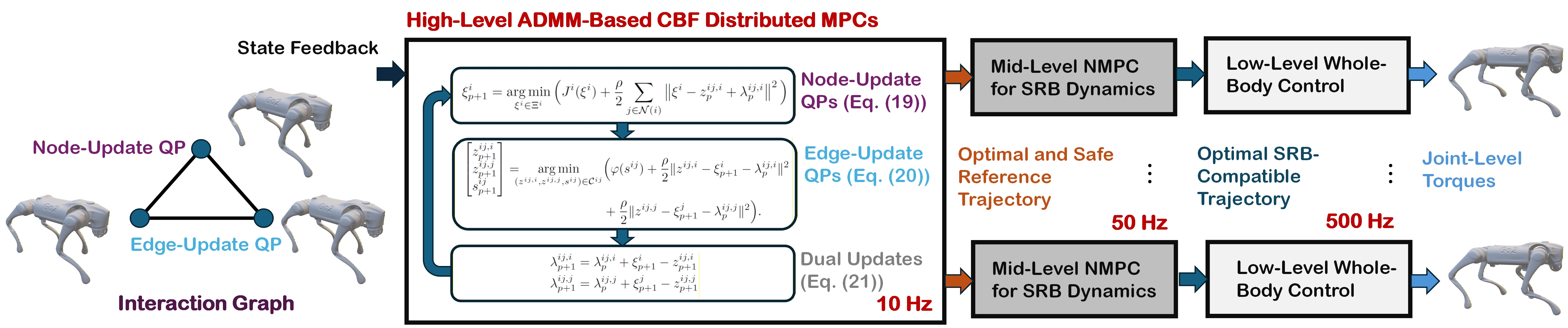}
    \vspace{-1.8em}
    \caption{Overview of the proposed layered control framework, consisting of a high-level ADMM-based CBF-DMPC for safety-critical trajectory planning, a mid-level NMPC layer enforcing the single rigid body (SRB) dynamics, and a low-level whole-body controller (WBC) enforcing full-order robot dynamics.}
    \vspace{-1.7em}
    \label{fig:Overview}
\end{figure*}


For multi-robot legged systems, distributed MPC (DMPC) provides an attractive framework by enabling each agent to compute its own trajectory while coordinating with neighboring agents through local communication. This decentralized structure reduces computational burden compared to centralized approaches, while enhancing robustness to communication failures and enabling real-time implementation.

Ensuring safety in multi-robot systems is critical, particularly in environments with static obstacles and close-proximity interactions among agents. Control barrier functions (CBFs) provide a principled framework for enforcing safety constraints by ensuring forward invariance of safe sets \cite{CBF_MRS,MRS_CBF_Cavorsi}. The integration of CBFs into MPC formulations enables safety-critical trajectory planning by explicitly incorporating safety constraints into the optimal control problem, while preserving performance and feasibility. This approach has been successfully applied to both single-robot and multi-robot quadrupedal systems \cite{NMPC_CBF_Sreenath,NMPC_CBF_Ames_Hutter,Basit_RAL}.

The incorporation of CBF constraints in MPC formulations for multi-robot legged systems introduces explicit inter-agent coupling, since the safety conditions depend on the states and inputs of neighboring agents. In distributed MPC formulations with embedded CBF constraints, this coupling remains embedded within the resulting nonlinear programming problems solved locally by each agent, see, e.g., \cite{Basit_RAL}. As a result, the optimization problems remain implicitly coupled and do not admit an explicit distributed optimization structure that separates node-local and edge-local computations through consensus constraints. This lack of structured decomposition limits the ability to fully exploit parallel computation and achieve decentralized implementations with symmetric computational load across agents. These limitations motivate the development of distributed MPC formulations with explicit consensus-based decomposition that enable fully decentralized trajectory optimization.

To address these limitations, the \textit{overarching goal} of this paper is to develop a fully decentralized MPC framework with CBF-based safety guarantees for teams of quadrupedal robots. To achieve this goal, we reformulate the centralized safety-critical MPC problem using a structured distributed optimization approach based on the alternating direction method of multipliers (ADMM). By introducing a node–edge splitting technique with consensus constraints, the proposed formulation decomposes the global problem into independent node-local and edge-local subproblems that can be solved in parallel using only neighbor-to-neighbor communication. This enables efficient real-time trajectory optimization with symmetric computational load across agents, while preserving safety and dynamic feasibility.


\vspace{-0.3em}
\subsection{Related Work}
\label{sec:Related_work}

Model predictive control MPC strategies for multi-robot systems can be broadly classified based on the level of coordination and information exchange among agents. Centralized MPC computes control inputs for all agents by solving a single global optimization problem, which can achieve optimal coordination but suffers from poor scalability due to the rapid growth in computational complexity and communication requirements as the number of agents increases \cite{Jeeseop_TRO}. At the opposite end of the spectrum, decentralized MPC assigns each agent an independent controller that operates without coordination, reducing computational burden but potentially resulting in suboptimal performance when inter-agent constraints such as collision avoidance are present \cite{Siljak_Decentralized_Book}.

Distributed MPC provides an intermediate solution by enabling agents to solve local optimization problems while exchanging information with neighboring agents to account for coupling effects \cite{2009_Scattolini_DMPC}. Depending on the coordination strategy, distributed MPC schemes may optimize a shared team-level objective through cooperative formulations \cite{venkat2005stability, rawlings2008coordinating}, or focus on individual agent objectives while incorporating inter-agent constraints through local coordination mechanisms \cite{richards2007robust, camponogara2002distributed}. These distributed approaches have been successfully applied to multi-agent coordination problems, including autonomous vehicle motion planning \cite{Abdelaal2019}, mobile robot navigation \cite{non-convex-PB-DMPC}, and networked dynamical systems \cite{FARINA20121088}. However, many distributed MPC formulations do not explicitly exploit structured distributed optimization techniques that enable parallelizable decomposition.

The ADMM is a widely used algorithm for distributed optimization, combining the decomposability of dual ascent methods with the convergence properties of the method of multipliers \cite{ADMM_Boyd}. While ADMM was originally developed for convex optimization problems, it has been extended to certain classes of nonconvex problems under appropriate assumptions \cite{ADMM_Nonconvex_Wang}. Nevertheless, extending ADMM to distributed MPC formulations with CBF constraints is not immediate, as the resulting optimization problems involve safety-critical constraints and explicit inter-agent coupling across the prediction horizon. These characteristics obscure the underlying separable structure and prevent direct application of standard ADMM, thereby requiring specialized reformulation techniques to expose a distributed optimization structure suitable for fully decentralized trajectory optimization.


\vspace{-0.3em}
\subsection{Contributions}
\label{sec:Contributions}

The main contributions of this paper are summarized as follows. We propose a novel distributed MPC framework with CBF constraints for safety-critical trajectory planning in multi-robot legged systems. Unlike prior distributed MPC formulations with CBF constraints \cite{Basit_RAL,Basit_ASME}, where inter-agent safety conditions remain embedded within coupled nonlinear programs solved locally by each agent, the proposed approach explicitly reformulates the centralized safety-critical MPC problem using a structured distributed optimization framework based on ADMM. By introducing a node–edge splitting technique with consensus constraints, this reformulation decomposes the centralized problem into independent node-local and edge-local subproblems, thereby enabling fully decentralized trajectory optimization with symmetric computational load across agents.

We derive distributed quadratic programming (QP) subproblems that can be solved in parallel using only local communication between neighboring agents. This explicit distributed formulation enables efficient real-time implementation. We integrate the proposed ADMM-based DMPC framework into a hierarchical locomotion control architecture for quadrupedal robots, consisting of high-level distributed trajectory planning, mid-level NMPC enforcing reduced-order SRB dynamics, and low-level nonlinear whole-body control enforcing full-order robot dynamics (see Fig. \ref{fig:Overview}).

We validate the proposed framework through hardware experiments on two Unitree Go2 robots and simulations with up to four agents navigating environments with rough terrain and disturbances (see Fig.~\ref{fig:OneSanpshot}). The distributed formulation achieves trajectory quality and safety comparable to centralized CBF-MPC, while reducing the average per-cycle planning time by approximately 51\% in the four-agent case, demonstrating improved computational efficiency for real-time decentralized implementation.


\vspace{-0.3em}
\section{Problem Formulation}
\label{sec:problem_formulation}

We consider a network of $n_{\Vset}$ quadrupedal robots indexed by the set $\Vset := \{1,\cdots,n_{\Vset}\}$. The superscript $i$ denotes quantities associated with agent $i \in \Vset$. The multi-robot system is modeled by an undirected interaction graph $\Ggraph(\Vset,\Eset)$, where 
$\Eset := \{\{i,j\} \mid i,j \in \Vset,\, i \neq j\}$ denotes the edge set. The discrete-time nonlinear dynamics of agent $i \in \Vset$ are given by
\begin{equation}\label{eq:local_dyn}
x^{i}(t+1)=f^{i}\big(x^{i}(t),u^{i}(t)\big),
\end{equation}
where $t \in \Integer := \{0,1,\cdots\}$ denotes the discrete-time index, 
$x^{i}(t) \in \Xset \subset \Real^{n_x}$ is the \textit{local state}, 
$u^{i}(t) \in \Uset \subset \Real^{n_u}$ is the \textit{local input}, 
and $f^{i} : \Xset \times \Uset \rightarrow \Xset$ 
represents the nonlinear dynamics of agent $i$. Let $g : \Xset \rightarrow \Real^{2}$ be a continuous mapping such that 
$g(x^{i})$ gives the Cartesian center-of-mass (CoM) position of agent $i$.

We consider a set of static obstacles indexed by 
$\Oset := \{1,\cdots,n_{\Oset}\}$, where $o^{\ell} \in \Real^{2}$ denotes the Cartesian center position of obstacle $\ell \in \Oset$. Our objective is to design a real-time distributed MPC framework based on ADMM and CBFs to steer the agents toward their goal configurations while guaranteeing safety with respect to both inter-agent and agent-obstacle collisions. For compact notation, define the \textit{global} state and input vectors as
\begin{equation}
x := \col\{x^{i}\}_{i\in\Vset} 
\in \Real^{n_{\Vset} n_x}, 
\quad
u := \col\{u^{i}\}_{i\in\Vset} 
\in \Real^{n_{\Vset} n_u},
\end{equation}
where ``\col'' is the column operator. To formally encode safety, we define the \textit{inter-agent safe set} as
\begin{equation*}
\mathcal{S}_{\Vset} 
:= \Big\{ x \in \Real^{n_{\Vset} n_x} \mid 
\|g(x^{i}) - g(x^{j})\| \ge d_{\thr}, 
\ \forall \{i,j\} \in \Eset \Big\},
\end{equation*}
and the \textit{agent-obstacle safe set} as
\begin{equation*}
\mathcal{S}_{\Oset} 
:= \Big\{ x \in \Real^{n_{\Vset} n_x} \mid 
\|g(x^{i}) - o^{\ell}\| \ge d_{\thr}, 
\ \forall i \in \Vset,\ \forall \ell \in \Oset \Big\},
\end{equation*}
where $\|\cdot\|$ denotes the Euclidean norm and $d_{\thr} > 0$ is a prescribed safety margin. The \textit{global safe set} is defined as
\begin{equation}
\mathcal{S} := \mathcal{S}_{\Vset} \cap \mathcal{S}_{\Oset}.
\end{equation}

To establish the proposed ADMM-based DMPC algorithm, we briefly review the concept of discrete-time CBFs.

\begin{definition}[Discrete-Time CBF {\cite{DT-HOCBF}}]
\label{def_CBF}
The local function $h^{i}$ is said to be a CBF for \eqref{eq:local_dyn} if there exists class $\mathcal{K}$ function $\alpha$ satisfying $\alpha(s) < s$ for all $s > 0$ such that 
\begin{equation}\label{eq:CBF_condition} 
\Delta h^{i}\left(x^{i}(t), u^{i}(t)\right) \geq - \alpha\left(h^{i}(x^{i}(t))\right), \quad \forall x^{i}(t) \in \mathcal{X}, 
\end{equation} 
where $\Delta h^{i}(x^{i}(t), u^{i}(t)) := h^{i}(x^{i}(t+1)) - h^{i}(x^{i}(t))$ and $s$ is the argument of $\alpha$.
\end{definition}

\begin{theorem} \textit{(CBF Condition \cite{DT-HOCBF}):}\label{Thm_CBF}
If $h^{i}$ is a continuous CBF, any control input $u^{i}(t)\in\mathcal{U}$ satisfying the CBF condition \eqref{eq:CBF_condition} will render the safety set given by $\Sset:=\{x^{i}:\,h^{i}(x^{i})\geq0\}$ forward invariant for agent $i\in\Vset$. 
\end{theorem}

To enforce safety for the multi-robot legged system, we introduce two classes of CBFs corresponding to agent-obstacle and inter-agent collision avoidance. Specifically, for each agent $i \in \Vset$, obstacle $\ell \in \Oset$, and edge $\{i,j\} \in \Eset$, define
\begin{align}
h^{i,\ell}(x^{i}) &:= \|g(x^{i}) - o^{\ell}\| - d_{\thr} \label{eq:CBF_obstacle}, \\
h^{i,j}(x^{i},x^{j}) &:= \|g(x^{i}) - g(x^{j})\| - d_{\thr}.\label{eq:CBF_inter_agent}
\end{align}
The corresponding CBF conditions take the form
\begin{align}
\Delta h^{i,\ell}(x^{i}(t),u^{i}(t)) 
&\geq -\alpha\big(h^{i,\ell}(x^{i}(t))\big)\label{eq:CBF_condition_obtscale}, \\
\Delta h^{i,j}(x^{i}(t),x^{j}(t),u^{i}(t),u^{j}(t)) 
&\geq -\alpha\big(h^{i,j}(x^{i}(t),x^{j}(t))\big)\label{eq:CBF_condition_inter_agent},
\end{align}
for all $i \in \Vset$, $\ell \in \Oset$, and $\{i,j\} \in \Eset$, where $\Delta h^{i,\ell}(x^{i}(t),u^{i}(t)):=h^{i,\ell}(x^{i}(t+1))-h^{i,\ell}(x^{i}(t))$ and $\Delta h^{i,j}(x^{i}(t),x^{j}(t),u^{i}(t),u^{j}(t)):=h^{i,j}(x^{i}(t+1),x^{j}(t+1))-h^{i,j}(x^{i}(t),x^{j}(t))$. 

From \eqref{eq:CBF_condition_obtscale}, the CBF condition for obstacle avoidance can be defined solely in terms of the state and control input of the local agent. However, as observed, the CBF condition \eqref{eq:CBF_condition_inter_agent} couples the states and inputs of the neighboring agents $i$ and $j$, which complicates the design of local MPCs for inter-agent collision avoidance. In what follows, we formulate the centralized MPC problem, which will be fully decentralized using the ADMM approach in Section \ref{sec:ADMM-based CBF-DMPC}. 

\textbf{Centralized MPC Problem:} Let us define the predicted local state and input trajectories at time $t$ over a control horizon of $N$ for agent $i \in \Vset$ by $x^{i}(\cdot) := \col\{x^{i}_{t+1|t}, x^{i}_{t+2|t}, \cdots, x^{i}_{t+N|t}\}\in\Real^{Nn_{x}}$ and $u^{i}(\cdot) := \col\{u^{i}_{t|t}, \cdots, u^{i}_{t+N-1|t}\}\in\Real^{Nn_{u}}$,
respectively, where $x^{i}_{t+k|t}$ and $u^{i}_{t+k|t}$ denote the predicted local state and local input at time $t+k$, computed at time $t$. The initial condition for the state prediction is taken as the actual local state, i.e., $x^{i}_{t|t} = x^{i}(t)$. 

By defining the \textit{state-input trajectory} for agent $i \in \Vset$ as $\xi^{i} := \col\big(x^{i}(\cdot), u^{i}(\cdot)\big)\in\Real^{N(n_{x}+n_{u})}$, the centralized CBF-based MPC problem can be expressed as
\begin{alignat}{4}
& \min_{(\xi^{1},\xi^{2},\cdots,\xi^{n_{\Vset}})} 
&& \sum_{i\in\Vset} J^{i}(\xi^{i}) \nonumber\\
& \quad \textrm{s.t.} 
&& x^{i}_{t+k+1|t}
= f^{i}(x^{i}_{t+k|t},u^{i}_{t+k|t}), \ i\in\Vset \nonumber\\
& && \Delta h^{i,\ell}_{t+k|t} 
\geq -\alpha\!\left( h^{i,\ell}_{t+k|t}\right), 
\quad i\in\Vset,\ \ell\in\Oset,\ \nonumber\\
& && \Delta h^{i,j}_{t+k|t} 
\geq -\alpha\!\left(h^{i,j}_{t+k|t}\right), 
\quad \{i,j\}\in\Eset,\ \nonumber\\
& && x^{i}_{t|t}=x^{i}(t), 
\quad i\in\Vset, \label{eq:centralized_MPC}
\end{alignat}
where
\begin{alignat}{4}
J^{i}(\xi^{i})
& := J^{i}_{\terminal}\!\left(x^{i}_{t+N|t}-x^{i,\des}_{t+N|t}\right) \nonumber\\
& \quad + \sum_{k=0}^{N-1} 
J^{i}_{\stage}\!\left(
x^{i}_{t+k|t}-x^{i,\des}_{t+k|t},
u^{i}_{t+k|t}
\right)
\end{alignat}
is a convex cost penalizing the deviation of the state trajectory $x^{i}(\cdot)$ from a given reference trajectory $x^{i,\des}(\cdot)$ using quadratic terminal and stage costs, namely,
\[
J^{i}_{\terminal}(x^{i}) = \|x^{i}\|^{2}_{P^{i}}, 
\qquad
J^{i}_{\stage}(x^{i},u^{i}) 
:= \|x^{i}\|^{2}_{Q^{i}} + \|u^{i}\|^{2}_{R^{i}},
\]
where $(P^{i}, Q^{i}, R^{i})$ are positive definite weighting matrices.

\textbf{Problem Statement:} 
We aim to design a computationally efficient, real-time distributed MPC algorithm based on the ADMM approach that computes a feasible and optimal local state-input trajectory for each agent $i \in \Vset$, denoted by $\xi^{i}$, while ensuring that the local system dynamics, the local CBF conditions for obstacle collision avoidance, and the inter-agent CBF conditions for robot-robot collision avoidance are satisfied.


\vspace{-0.3em}
\section{ADMM-based DMPC with CBF Constraints}
\label{sec:ADMM-based CBF-DMPC}

This section presents the proposed high-level ADMM-based CBF-DMPC framework for real-time trajectory planning and safe locomotion of quadrupedal robots in complex environments. 

As discussed earlier, the dynamics \eqref{eq:local_dyn} and the obstacle-avoidance CBF condition \eqref{eq:CBF_condition_obtscale} are defined locally. Let us define the \textit{local feasible set} for the trajectory $\xi^{i}$, compatible with the local dynamics and the obstacle-avoidance CBF condition, as
\begin{alignat}{4}
    &\Xi^{i}:=\Big\{\xi^{i} \mid\, && \textrm{Local dynamics \eqref{eq:local_dyn} and the local CBF condition} \nonumber\\
    & && \textrm{\eqref{eq:CBF_condition_obtscale} are satisfied}\Big\}.
\end{alignat}
Then, the centralized MPC problem \eqref{eq:centralized_MPC} can be rewritten as follows:
\begin{alignat}{4}
& \min_{\{\xi^{i}\}} 
&& \sum_{i\in\Vset} J^{i}(\xi^{i}) && \nonumber\\
& \quad \textrm{s.t.} \quad 
&& \xi^{i}\in\Xi^{i},\quad i\in\Vset, && (\textrm{Local Feasibility})\nonumber\\
& && \psi^{ij}(\xi^{i},\xi^{j})\leq0,\quad \{i,j\}\in\Eset,\,\,  && (\textrm{Coupled Inequity}) \label{eq:centralized_MPC_ver2}
\end{alignat}
where the inequality constraint $\psi^{ij}(\xi^{i},\xi^{j}) \leq 0$ represents the inter-agent CBF condition between the trajectories of agents $i$ and $j$.

In the centralized problem \eqref{eq:centralized_MPC_ver2}, the convex cost function is separable, subject to local feasibility constraints and coupled inequality constraints arising from the inter-agent CBF condition. In the original formulation of the ADMM approach \cite[Chap. 3]{ADMM_Boyd}, the coupling appears in equality constraints. Hence, we introduce the set of slack variables $\{s^{ij}\}$ with $\{i,j\}\in\Eset$ to reformulate the coupled inequality constraints as coupled equality constraints. Specifically, the new problem is given as follows:
\begin{alignat}{4}
& \min_{\{\xi^{i}\},\{s^{ij}\}} 
&& \sum_{i\in\Vset} J^{i}(\xi^{i}) + \sum_{\{i,j\}\in\Eset} \varphi(s^{ij})&& \nonumber\\
& \quad \textrm{s.t.}
&& \xi^{i}\in\Xi^{i},\quad i\in\Vset, && (\textrm{Local Feasibility})\nonumber\\
& && \psi^{ij}(\xi^{i},\xi^{j})+s^{ij}=0,\, \{i,j\}\in\Eset,\, && (\textrm{Coupled Equality})\nonumber\\
& && s^{ij}\geq0,\ \{i,j\}\in\Eset, && \label{eq:centralized_MPC_ver3}
\end{alignat}
where $\varphi(s^{ij})$ is a convex quadratic penalty on the slack variables to ensure that the overall cost function is positive definite with respect to the decision variables $\{\xi^{i}\}$ and $\{s^{ij}\}$. 


\vspace{-0.3em}
\subsection{Proposed ADMM-based DMPC with CBFs}
\label{sec:Proposed ADMM-based DMPC}

To obtain a \textit{fully decentralized ADMM} formulation, we propose a \textit{node-edge splitting} technique by keeping the true decision variable $\xi^{i}$ at node $i$, and creating \textit{edge-local copies} of the variables that appear in each pairwise constraint. Then, ADMM enforces \textit{consensus} between the node variables and the corresponding edge copies. This removes all direct coupling from the node subproblems, and the only remaining coupling appears in small edge subproblems that involve only local ${i,j}$ information. In what follows, we describe the procedure in detail.

For every $\{i,j\}\in\Eset$, we introduce the \textit{edge variables} 
\begin{equation}
    z^{ij,i} \approx \xi^{i}, \quad z^{ij,j} \approx \xi^{j}, \quad s^{ij}\geq0,
\end{equation}
and impose the \textit{consensus constraints}
\begin{equation}
    z^{ij,i}=\xi^{i}, \quad z^{ij,j} = \xi^{j}, \quad \forall \{i,j\}\in\Eset. 
\end{equation}
We also enforce the original edge feasibility locally at each edge:
\begin{equation}
    \psi^{ij}(z^{ij,i},z^{ij,j})+s^{ij}=0,\quad s^{ij}\geq0. 
\end{equation}
Then, the problem \eqref{eq:centralized_MPC_ver3} becomes
\begin{alignat}{4}
    &\min_{\{\xi^{i}\},\{z^{ij,i},z^{ij,j},s^{ij}\}} &&\sum_{i\in\Vset} \left(J^{i}(\xi^{i}) + \mathbb{I}_{\Xi^{i}}(\xi^{i})\right) \nonumber\\
    & &&+ \sum_{\{i,j\}\in\Eset} \left(\varphi(s^{ij}) + \mathbb{I}_{\Cset^{ij}}(z^{ij,i},z^{ij,j},s^{ij})\right)\nonumber\\
    & && \textrm{s.t.}\,\,\, z^{ij,i}=\xi^{i},\quad z^{ij,j}=\xi^{j},
\end{alignat}
where 
\begin{equation}
    \Cset^{ij}:=\{(z^{ij,i},z^{ij,j},s^{ij}) \mid \psi^{ij}(z^{ij,i},z^{ij,j})+s^{ij}=0, s^{ij}>0\},
\end{equation}
and $\mathbb{I}_{\Xi^{i}}$ and $\mathbb{I}_{\Cset^{ij}}$ denote the indicator functions corresponding to the sets $\Xi^{i}$ and $\Cset^{ij}$, respectively, i.e.,
\[
\mathbb{I}_{\Xi^{i}}(\xi^{i}) :=
\begin{cases}
0, & \text{if } \xi^{i} \in \Xi^{i}, \\
+\infty, & \text{if } \xi^{i} \notin \Xi^{i}.
\end{cases}
\]
This formulation is now in the standard ADMM form, consisting of a sum of node-local terms, a sum of edge-local terms, and consensus constraints. Therefore, the ADMM approach can be directly applied.

\textbf{Scaled ADMM Iterations:} We are now in a position to present the steps of the proposed ADMM approach. The algorithm consists of three steps: 1) local node updates at each agent $i \in \Vset$, 2) edge updates for each edge $\{i,j\} \in \Eset$, and 3) dual variable updates. In our notation, the subscript $p$ denotes the iteration index of the ADMM algorithm. Suppose further that $\lambda^{ij,i}$ and $\lambda^{ij,j}$ denote the \textit{scaled Lagrange (i.e., dual) variables} corresponding to the equality constraints $z^{ij,i}=\xi^{i}$ and $z^{ij,j}=\xi^{j}$, respectively.

\textit{Step 1: Node update (at each agent $i\in\Vset$, local):} 
During iteration $p+1$, we update the local trajectory for agent $i$ using $(z^{ij,i}_{p}, \lambda^{ij,i}_{p})$, computed at iteration $p$, as follows:
\begin{equation}\label{eq:node_update}
    \xi^{i}_{p+1} = \argmin_{\xi^{i}\in\Xi^{i}} 
    \Big(J^{i}(\xi^{i}) +
    \frac{\rho}{2} \sum_{j\in\Nset(i)} \left\|
    \xi^{i} - z^{ij,i}_{p} + \lambda^{ij,i}_{p}
    \right\|^{2}\Big),
\end{equation}
where $\Nset(i)$ denotes the set of neighboring agents of agent $i$ and $\rho>0$ is the penalty parameter. The node update optimization is performed in parallel across all agents.

\textit{Step 2: Edge update (for each $\{i,j\}\in\Eset$):} After completing the node updates, we solve the following optimization problem to update edge variables $(z^{ij,i}_{p+1}, z^{ij,j}_{p+1}, s^{ij}_{p+1})$ using $(\xi^{i}_{p+1}, \xi^{j}_{p+1})$:
\begin{alignat}{4}
    &\begin{bmatrix}
        z^{ij,i}_{p+1}\\
        z^{ij,j}_{p+1}\\
        s^{ij}_{p+1}
    \end{bmatrix}=\!\!\!\!\argmin_{(z^{ij,i},z^{ij,j},s^{ij})\in\Cset^{ij}} && \Big( \varphi(s^{ij}) + \frac{\rho}{2}\|z^{ij,i} - \xi^{i}_{p+1} - \lambda^{ij,i}_{p}\|^{2} \nonumber\\
    & &&+ \frac{\rho}{2}\|z^{ij,j} - \xi^{j}_{p+1} - \lambda^{ij,j}_{p}\|^{2}\Big).\label{eq:edge_update}
\end{alignat}
This edge update requires only the current messages $(\xi^{i}_{p+1}, \lambda^{ij,i}_{p})$ from node $i$ and $(\xi^{j}_{p+1}, \lambda^{ij,j}_{p})$ from node $j$.

\textit{Step 3: Dual updates (at the endpoints):} Using the updated values $(\xi^{i}_{p+1}, \xi^{j}_{p+1}, z^{ij,i}_{p+1}, z^{ij,j}_{p+1})$, the dual variables are updated as
\begin{alignat}{4}
    &\lambda^{ij,i}_{p+1} && = \lambda^{ij,i}_{p} + \xi^{i}_{p+1} - z^{ij,i}_{p+1} \nonumber 
    \\
    &\lambda^{ij,j}_{p+1} && = \lambda^{ij,j}_{p} + \xi^{j}_{p+1} - z^{ij,j}_{p+1}. 
\end{alignat}

\textit{Communication Protocol:} After each iteration, each agent $i$ exchanges information with each neighbor $j \in \Nset(i)$ as follows: (a) agent $i$ sends its updated trajectory $\xi^{i}_{p+1}$ to agent $j$, and (b) after solving the edge subproblem, the agents exchange the resulting edge variables $z^{ij,i}_{p+1}$ and $z^{ij,j}_{p+1}$. In the proposed approach, all computations and communications are localized to nodes and edges.

\textit{Computational Load:} At each ADMM iteration, the proposed decomposition of the CBF-based DMPC results in two sets of independent optimization problems. First, each agent $i$ performs one node update by solving the local optimization problem \eqref{eq:node_update} with respect to $\xi^{i}$, resulting in a total of $n_{\Vset}=|\Vset|$ node-local problems. Second, each edge $(i,j)\in\mathcal{E}$ performs one edge update as in \eqref{eq:edge_update} to compute the corresponding auxiliary variables, resulting in $n_{\Eset}:=|\mathcal{E}|$ edge-local problems. All node and edge problems are mutually independent and can be solved in parallel using only local information exchange between neighboring agents. Consequently, each iteration consists of $n_{\Vset} + n_{\Eset}$ decentralized optimization problems, and for regular graphs such as the complete graph, the computational load is distributed symmetrically across agents.


\vspace{-0.3em}
\subsection{Reduction to Distributed QPs and Layered Control}
\label{sec:Reduction to Distributed QPs and Layered Control}

This section provides a concise overview of the high-, mid-, and low-level layers of the proposed layered control scheme for teams of quadrupedal robots.

At the high level of the layered control structure, we employ the proposed ADMM-based DMPC with CBFs on a team of agents modeled using the kinematic car model, given by
\begin{equation}\label{eq:kinematic_car}
    \begin{bmatrix}
    \dot{p}_{x}^{i}\\
    \dot{p}_{y}^{i}\\
    \dot{\theta}^{i}
    \end{bmatrix}
    =
    \begin{bmatrix}
        v^{i}\cos(\theta^{i})\\
        v^{i}\sin(\theta^{i})\\
        \omega^{i}
    \end{bmatrix},
\end{equation}
where $x^{i} := \col(p_{x}^{i},p_{y}^{i},\theta^{i}) \in \Real^{3}$ denotes the local state vector of agent $i$, with $p_{x}^{i}$ and $p_{y}^{i}$ representing the Cartesian position of robot $i$ in the $xy$-plane and $\theta^{i}$ denoting the yaw angle. The local control input for agent $i$ is defined as $u^{i} := \col(v^{i},\omega^{i}) \in \Real^{2}$, where $v^{i}$ and $\omega^{i}$ denote the linear and angular velocities, respectively. The continuous-time kinematic car model in \eqref{eq:kinematic_car} is discretized using the Euler method for the DMPC implementation. 

In this work, we further successively linearize the discrete-time dynamics of each agent and the CBF conditions \eqref{eq:CBF_condition_obtscale} and \eqref{eq:CBF_condition_inter_agent} around the current operating point. This transforms the node and edge subproblems into a set of decentralized QPs that can be solved efficiently in parallel. Section~\ref{sec:Setup and Controller Synthesis} provides the computational details of the proposed ADMM-based distributed QP formulation with CBF constraints.


\begin{figure*}[t]
    \centering
    \includegraphics[width=1.00\linewidth]{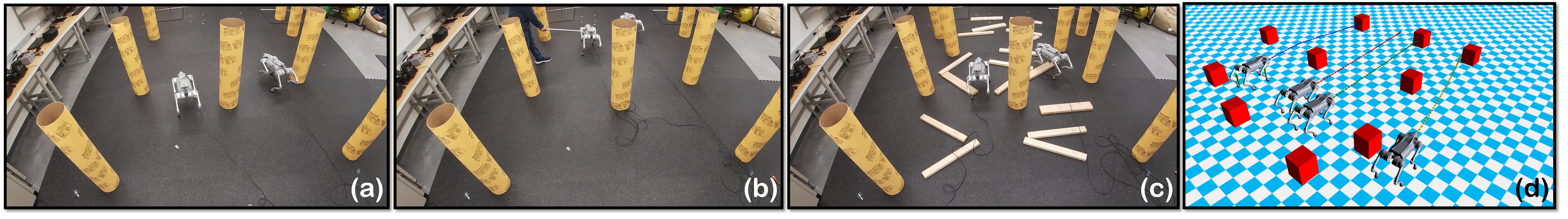}
    \vspace{-2em}
    \caption{Hardware experiments of the proposed ADMM-based CBF-DMPC and layered control with two Unitree Go2 robots: (a) navigation among eight static obstacles (experiment 1), (b) disturbance rejection under external pushes (experiment 2), and (c) obstacle avoidance over rough terrain with ten obstacles (experiment 3). (d) Multi-agent simulation in RaiSim with four robots navigating among ten obstacles.}
    \vspace{-1.1em} 
    \label{fig:MultipleSanpshots}
\end{figure*}


At the mid level, we employ a 50 Hz nonlinear MPC (NMPC) that enforces the reduced-order single rigid body (SRB) dynamics of each quadruped to track the high-level ADMM-based CBF-DMPC references. The NMPC incorporates the full 6-DoF SRB dynamics under ground reaction forces (GRFs) \cite{Basit_RAL,Ruturaj_Locomanipulation_RAL} and optimally computes GRFs to ensure dynamic feasibility, stability, friction-cone compliance, and accurate velocity tracking.

At the low level, a nonlinear whole-body controller (WBC) enforces the full-order robot dynamics to track the desired SRB states and GRFs generated by the NMPC. The WBC, adopted from \cite{Randy_Paper_LCSS,Ruturaj_Locomanipulation_RAL}, is formulated as a 500 Hz real-time QP that respects full-body dynamics and actuation constraints.


\vspace{-0.3em}
\section{Experiments}
\label{sec:Experiments}

This section presents the results of our numerical simulations and hardware experiments with the proposed layered control scheme. 

\vspace{-0.3em}
\subsection{Setup and Controller Synthesis}
\label{sec:Setup and Controller Synthesis}

This work employs Unitree Go2 quadrupedal robots for both numerical simulations and hardware experiments. Each Go2 robot has a mass of 15~kg, a nominal height of approximately 0.28~m, and 18 degrees of freedom (DoFs), including 12 actuated leg joints (hip roll, hip pitch, and knee pitch for each leg) and 6 unactuated floating-base DoFs describing the position and orientation of the torso. Each robot is equipped with a Unitree L1 4D LiDAR sensor for environment perception and localization.

The proposed layered control architecture is implemented using a multi-threaded framework on an offboard desktop computer equipped with an Intel i9-12900K CPU and 128~GB of DDR5 RAM. The high-level ADMM-based CBF-DMPC is executed in a fully decentralized manner, where node and edge update problems are solved in parallel across independent threads at 10~Hz. In addition, separate threads execute the mid-level nonlinear MPC controllers at 50~Hz and the low-level whole-body controllers at 500~Hz. This parallelized implementation enables real-time execution of the distributed optimization and control framework while maintaining computational efficiency. The translational motion of each robot is estimated using a kinematic state estimator. Numerical simulations are conducted in the RaiSim physics engine~\cite{RAISIM} using up to four Go2 robots, while hardware experiments are performed using two physical Go2 platforms. Supplementary videos demonstrating the experimental and numerical results are available with the submission.



\begin{table}[t]
\centering
\caption{Statistics of QP computation times (ms).}
\label{tab:qp_timing}
\vspace{-1em}
\begin{tabular}{c|cc|cc}
\hline
\multirow{2}{*}{QP Type} 
& \multicolumn{2}{c|}{Two Agents} 
& \multicolumn{2}{c}{Four Agents } \\
&  Avg & Std & Avg & Std \\
\hline
Node-update QP &  0.51 & 0.71  & 0.92 & 8.34 \\
Edge-update QP &  0.002 & 0.000  & 0.009 & 0.003 \\
\textbf{Total (ADMM)} & \textbf{8.06} & \textbf{10.66}  & \textbf{14.58} & \textbf{11.95}\\
Total (Centralized)  & 9.61 & 3.87  & 29.94 & 21.75
\end{tabular}
\vspace{-2em}
\end{table}



\textit{Real-Time Computation:} The weighting matrices of the ADMM-based CBF-DMPC are selected as $Q^{i}=\diag\{50,50,100\}$, $R^{i}=\diag\{50, 10\}$, and $P^{i}=10\,Q^{i}$ for all agents $i \in \Vset$ with $\varphi(s)=5s^{2}$. The prediction horizon of the local MPC is chosen as $N=50$, with a sampling time of $T_{s}=0.1$~s. The safety and ADMM hyperparameters are chosen as follows. The collision avoidance threshold distance is set to $d_{\thr}=0.5$~m. The class $\mathcal{K}$ function used in the CBF conditions is selected as $\alpha(s)=0.3\,s$. The ADMM penalty parameter is set to $\rho=20$ to balance convergence speed and numerical stability in the node and edge updates. The maximum number of ADMM iterations per control cycle is limited to 15, enabling real-time execution while ensuring convergence of the distributed optimization. We choose trotting gaits for the quadrupedal robots. For the two- and four-agent simulations, we have $n_{\mathcal{V}}=2$ and $n_{\mathcal{V}}=4$ node-update QPs, respectively, in each ADMM iteration. Each of these QPs consists of $(3+2)N=250$ decision variables corresponding to the local trajectory $\xi^{i}$. For the two- and four-agent simulations, we have $n_{\mathcal{E}}=1$ and $n_{\mathcal{E}}=6$ edge-update QPs, respectively, in each ADMM iteration. Each of these edge-update QPs consists of $2\times(3+2)N+1=501$ decision variables corresponding to the edge variables and the associated slack variable. The local QPs are solved using the OSQP solver~\cite{osqp} with a maximum of 200 iterations. The mid-level NMPC is solved at 50~Hz using the CasADi framework \cite{CasADI} with the FATROP solver \cite{vanroye2023fatrop}, with a maximum of 12 iterations, similar to~\cite{Ruturaj_Locomanipulation_RAL}.

We evaluate the framework in a complex environment with up to $n_{\mathcal{O}}=10$ obstacles. Table~\ref{tab:qp_timing} summarizes the statistics of the QP computation times (in milliseconds) for the node-update and edge-update steps across ADMM iterations. It also reports the total per-cycle planning times for both the ADMM-based CBF-DMPC and the centralized CBF-MPC \eqref{eq:centralized_MPC_ver3} for the two- and four-agent cases (see Sec. \ref{sec:Comparison Analysis}). 


\vspace{-0.3em}
\subsection{Hardware Experiments}
\label{sec:Hardware Experiments}

We experimentally validate the proposed ADMM-based CBF-DMPC framework on two Go2 robots operating in an indoor laboratory environment (see Fig. \ref{fig:MultipleSanpshots}). The experiments are designed to evaluate (i) safety-critical navigation in cluttered environments, (ii) robustness to external disturbances, and (iii) performance over rough terrain. 

\textit{Experiment 1: Cluttered Environment with Eight Obstacles:} In the first experiment, eight static obstacles are distributed within the workspace. Each robot is assigned an individual goal configuration and commanded to follow its reference trajectory while avoiding both obstacles and the neighboring robot. The distributed CBF-DMPC successfully generates collision-free trajectories in real time. Throughout the experiment, the inter-robot distance remained above the prescribed safety threshold $d_{\mathrm{th}}$, and no obstacle collisions were observed (see Fig. \ref{fig:MultipleSanpshots}(a)). The robots autonomously adjust their trajectories through local edge updates without requiring centralized coordination.

\begin{figure}[t]
    \centering
    \includegraphics[width=\linewidth]{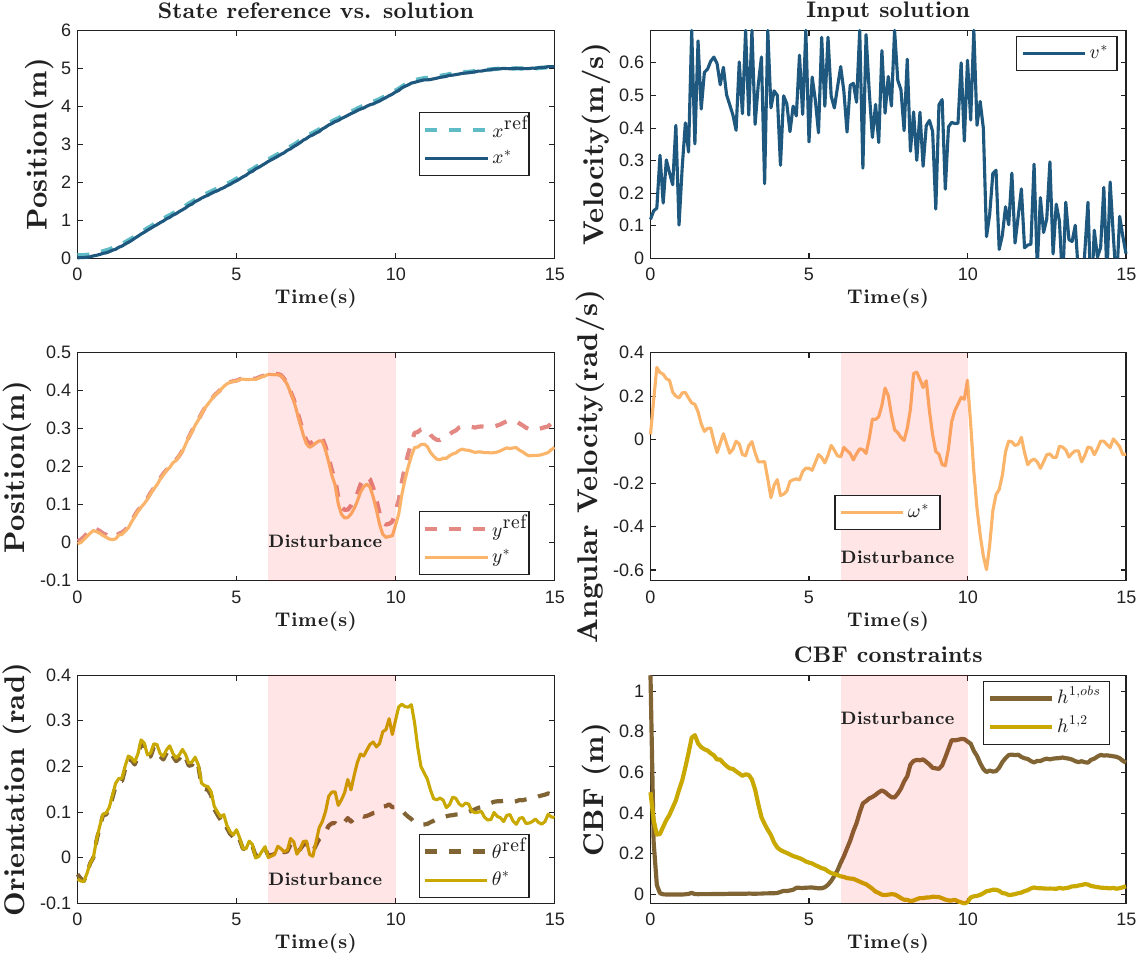}
    \vspace{-1.8em}
    \caption{Plot of the optimal state trajectory $(x^{\star,i},y^{\star,i},\theta^{\star,i})$ generated by the high-level ADMM-based CBF-DMPC for agent~1, together with the corresponding reference trajectory toward its goal during experiment 2. The figure also depicts the optimal velocity commands $(v^{\star,i},\omega^{\star,i})$ and the minimum CBF values, representing obstacle-avoidance and inter-agent safety margins. The shaded regions indicate push intervals.}
    \vspace{-1.3em} 
    \label{fig:Push_Experiment}
\end{figure}

\textit{Experiment 2: Robustness to External Disturbances:}
In the second experiment, a bystander physically perturbs one of the robots multiple times during locomotion to evaluate robustness (see Fig. \ref{fig:MultipleSanpshots}(b)). These pushes introduce significant deviations from the nominal trajectory and temporary mismatches between the predicted and actual robot motion. Despite these disturbances, the proposed layered control structure maintains safety and convergence toward the assigned goals. Figure~\ref{fig:Push_Experiment} illustrates the optimal state trajectory $(x^{i},y^{i},\theta^{i})$ generated by the high-level ADMM-based CBF-DMPC for agent~1, together with the corresponding reference trajectory toward its goal. The reference trajectory is generated using an interpolating B\'ezier polynomial connecting the current state to the goal configuration. The figure also presents the optimal velocity commands $(v^{i},\omega^{i})$ and the minimum CBF values $h^{i,\mathrm{obs}} := \min_{\ell \in \Oset} h^{i,\ell}$ and $h^{i,j}$, representing obstacle-avoidance and inter-agent safety margins, respectively. As observed, the CBF functions remain non-negative under nominal operation, ensuring forward invariance of the safe set. During the push intervals, temporary reductions in the CBF values occur due to disturbances that are not captured in the model. However, the proposed ADMM-based CBF-DMPC rapidly replans the trajectories and restores the CBF values to the non-negative region, thereby recovering safety. 


\textit{Experiment 3: Navigation on Rough Terrain with Ten Obstacles:} In the third experiment, ten obstacles are placed in the environment, and wooden blocks are randomly distributed on the laboratory floor to create rough terrain conditions (see Fig. \ref{fig:MultipleSanpshots}(c)). The robots are commanded to reach their individual goals while traversing uneven surfaces and avoiding both static obstacles and each other. This scenario tests the interaction between high-level distributed planning and low-level dynamic feasibility enforcement. The proposed framework maintains safe inter-agent distances and obstacle avoidance while the mid-level NMPC and whole-body controller ensure stable locomotion over uneven terrain. The experimental results demonstrate that the distributed safety-critical planning layer integrates effectively with dynamic locomotion control in challenging physical conditions.

\textit{Simulation with Four Agents:} While hardware validation is conducted on two physical robots, we additionally report successful numerical simulations involving four quadrupedal robots navigating in shared environments with random $n_{\Oset}=10$ obstacles (see Fig, \ref{fig:MultipleSanpshots}(d)). In the four-agent case, the interaction graph contains $n_{\Vset} = 4$ nodes and $n_{\Eset} = 6$ edges, resulting in four node-update QPs and six edge-update QPs per ADMM iteration. These simulations confirm that the proposed structured ADMM-based CBF-DMPC formulation extends naturally beyond two agents while preserving decentralized computation and safety constraints.


\vspace{-0.3em}
\subsection{Comparison Analysis}
\label{sec:Comparison Analysis}

We compare the total wall-clock computation time per sampling instant for the centralized CBF-MPC \eqref{eq:centralized_MPC_ver3} and the proposed ADMM-based CBF-DMPC in RaiSim for the two- and four-agent cases. Both approaches solve quadratic programs obtained from the same successive linearization procedure, using identical horizon length, cost weights, safety parameters, and solver configurations. In each control cycle, the centralized method solves a single coupled QP over the stacked decision variables of all agents, whereas the proposed approach performs node-update and edge-update QPs \eqref{eq:node_update} and \eqref{eq:edge_update} within the ADMM framework and returns the distributed solution after 15 ADMM iterations (see Table~\ref{tab:qp_timing}).

\begin{figure}[t]
    \centering
    \includegraphics[width=0.75\linewidth]{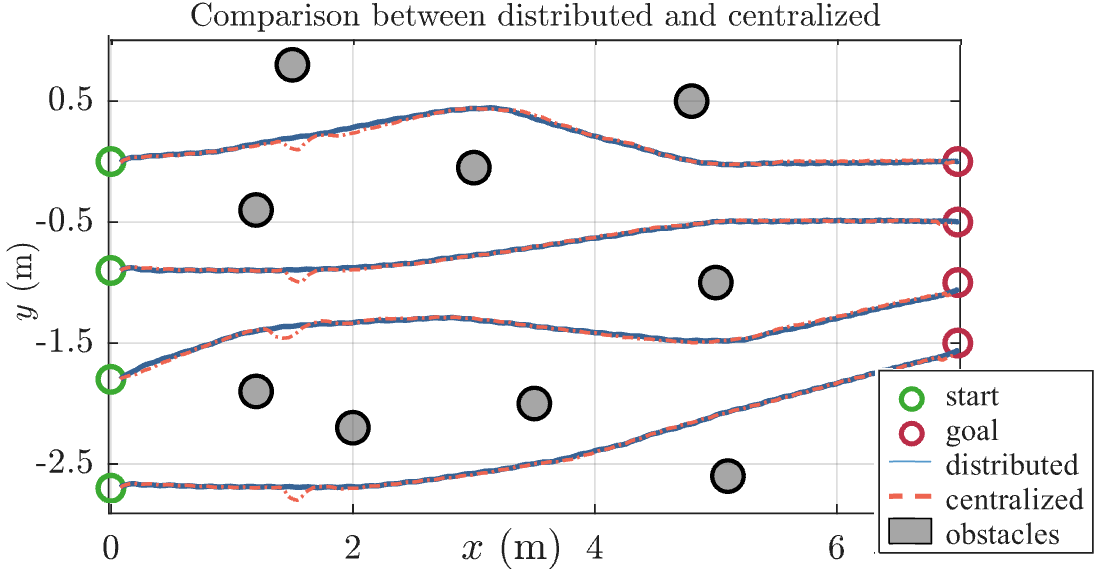}
    \vspace{-1.1em}
    \caption{Comparison between the ADMM-based CBF-DMPC and centralized CBF-MPC for four agents navigating among obstacles in simulation. The resulting $xy$ trajectories demonstrate near-identical motion and safety behavior under both approaches.}
    \vspace{-1.0em} 
    \label{fig:comparison}
\end{figure}

Over the full simulation horizon, the ADMM-based CBF-DMPC reduces the average per-cycle planning time by approximately $16\%$ and $51\%$ relative to the centralized CBF-MPC in the two- and four-agent cases, respectively. Moreover, the distributed solution closely matches the centralized optimal solution in terms of trajectory quality and safety margins, with negligible differences observed in the objective values and CBF constraint satisfaction (see Fig.~\ref{fig:comparison}). These results demonstrate that the proposed structured ADMM decomposition improves computational efficiency while preserving solution quality and safety, with increasing benefits as the number of agents grows.


\vspace{-0.3em}
\section{Conclusions}
\label{sec:Conclusions}

This paper presented a fully decentralized MPC framework with CBF constraints for safety-critical trajectory planning in multi-robot quadrupedal systems. By reformulating the centralized CBF-based MPC problem using a structured ADMM approach with a node–edge splitting technique and consensus constraints, the proposed method decomposes the coupled optimization problem into independent node-local and edge-local quadratic programs that can be solved in parallel using only neighbor-to-neighbor communication. This structured decomposition enables symmetric computational load across agents while preserving safety guarantees.


The proposed ADMM-based CBF-DMPC was integrated into a layered locomotion control architecture and validated through hardware experiments on two Unitree Go2 robots and numerical simulations with up to four agents in complex environments with obstacles, rough terrain, and external disturbances. The results demonstrate that the distributed formulation achieves performance comparable to centralized CBF-MPC while reducing the average per-cycle planning time by approximately 16\% and 51\% in the two- and four-agent cases, respectively. These findings indicate increasing computational benefits as the number of agents increases from two to four. Future work will investigate the scalability of the proposed framework to larger teams of quadrupedal robots and explore fully decentralized implementations using distributed onboard computing architectures. 




\vspace{-0.5em}
\bibliographystyle{IEEEtran}
\bibliography{references}

\end{document}